\documentclass[a4paper]{article}

\usepackage[english]{babel}
\usepackage[utf8x]{inputenc}
\usepackage[T1]{fontenc}

\usepackage[a4paper,top=3cm,bottom=2cm,left=3cm,right=3cm,marginparwidth=1.75cm]{geometry}

\usepackage{amsmath, amsfonts,amssymb,amsthm}
\usepackage{graphicx}
\usepackage[colorinlistoftodos]{todonotes}
\usepackage[colorlinks=true, allcolors=blue]{hyperref}

\usepackage{lineno,hyperref}
\usepackage{subfigure}
\usepackage{algorithmic}
\usepackage{float}
\usepackage[affil-it]{authblk}
\usepackage{cases}
\usepackage{multirow}

\title{Simultaneously Solving Mixed Model Assembly Line Balancing and Sequencing problems with FSS Algorithm}
\author[1]{J.B. Monteiro-Filho*}
\author[1]{I.M.C. Albuquerque}
\author[1]{F.B. Lima Neto}
\affil[1]{Computational intelligence research group - Polytechnical School of Pernambuco Benfica 455, Recife-PE, Brazil}
\affil[*]{Corresponding author: jbmf@ecomp.poli.br}

\begin{document}
\maketitle

\begin{abstract}
Many assembly lines related optimization problems have been tackled by researchers in the last decades due to its relevance for the decision makers within manufacturing industry. Many of theses problems, more specifically Assembly Lines Balancing and Sequencing problems, are known to be NP-Hard. Therefore, Computational Intelligence solution approaches have been conceived in order to provide practical use decision making tools. In this work, we proposed a simultaneous solution approach in order to tackle both Balancing and Sequencing problems utilizing an effective meta-heuristic algorithm referred as Fish School Search. Three different test instances were solved with the original and two modified versions of this algorithm and the results were compared with Particle Swarm Optimization Algorithm.
\end{abstract}

\section{Introduction}
\label{intro}

Products requirements and consequently the requirements of production systems have changed dramatically over the years. Back in 1913, when Henry Ford conceived the Assembly Lines (AL), this manufacturing paradigm was developed for the mass production of standardized products. However, nowadays increasing variety of models and options to be selected by customers have changed the market scenario and manufacturers have to handle a variety which frequently exceeds several billions of models \cite{Boysen2009}. As an example, the German car manufacturer BMW produces $10^{32}$ different car models \cite{Boysen2008}.

The high variety issue typically requires implementation of cost efficient flexible production systems. This is normally addressed by the use of the so called mixed-model assembly, where setup operations are reduced to such an extent that various models of a common base product can be manufactured in intermixed sequences \cite{Moradi2012}.

Mixed-model assembly lines (MMAL) facilitate product variations and diversities on the same line in an intermixed scenario. Hence, optimal AL design, balancing and product sequencing of mixed-models assembly are the major challenges for manufacturers for creating high-variety and low-volume production within the process \cite{Uddin2010}.

A novel Assembly Line Balancing Problem (ALBP) referred as Mixed Model Workplace Time Dependent Assembly Line Balancing Problem (MMWALBP) was recently proposed. Due to its mixed model feature, a Mixed Model Sequencing Problem arises (MMSP), as this is the case for any mixed model ALBP.

Mixed Model Sequencing and Assembly lines Balancing are strongly related problems. The sequencing problem depends on a given balancing solution in order to be solved and the quality of the sequencing depends on the quality of the balancing solution provided \cite{scholl1999balancing}. Both problems are known to be NP-Hard \cite{Becker2006}. Therefore, many different procedures have been developed in order to tackle the assembly line related problems \cite{Scholl2006}. Moreover, metaheuristic approaches have been playing an important role in their solutions as well as in many other classes of combinatorial optimization problems \cite{blum2003metaheuristics}.

The Fish School Search algorithm (FSS), presented originally in 2008 by Bastos-Filho and Lima Neto \cite{Filho2008}, is a population based continuous optimization technique inspired in the behavior of fish schools while looking for food. Each fish in the school represents a solution for a given optimization problem and the algorithm utilizes the information of every fish to regulate between exploration/exploitation behaviors and guide the school to promising regions in the search space as well as avoiding early convergence in local optima.

Variations of FSS were proposed and applied in the solution of ALBP \cite{albuquerque2016} \cite{albuquerque2016b}. In the present work, an integrated approach was employed to simultaneously solve MMWALBP and MMSP with the original version of FSS, referred as FSS vanilla (FSS-V), as well as with two variations of the algorithm called FSS Stagnation Avoidance Routine (FSS-SAR) \cite{monteiro2016} and FSS Not Penalizing Static Success (FSS-NPSS).

The remainder of this article is organized as follows: Section \ref{FSS} defines FSS-V, FSS-SAR and FSS-NPSS. Section \ref{problemDef} introduces the two problems tackled within this research effort. Section \ref{MMSPPFSS} presents the simultaneous solution procedure applied to solve both problems and also the specific approaches utilized to tackle each problem individually. Sections \ref{experiments} and \ref{conclusion} presents the tests performed and the results found.

\section{Fish School Search Algorithm}
\label{FSS}

\subsection{FSS - Vanilla}
\label{FSSV}

FSS is a population based search algorithm inspired in the behavior of a school of swimming fishes that expands and contracts while looking for food. Each fish $n$-dimensional location represents a possible solution for the optimization problem. The algorithm makes use of weights for all fishes which represents cumulative account on how successful has been the search for each fish in the school \cite{Filho2008}.

FSS is composed of four operators which run sequentially within each iteration. They are: Individual movement, feeding, Collective-instinctive movement and Collective-volitive movement.

The general pseudo-code for FSS is:

\begin{algorithmic}[1]
	\label{FSSPseudoCode}
	\STATE Initialize user parameters 
	\STATE Initialize fishes positions randomly 
	\WHILE{Stopping condition is not met}
	\STATE Calculate fitness for each fish 	
	\STATE Run Individual operator movement
	\STATE Run Feed operator 
	\STATE Run Collective-instinctive movement operator
	\STATE Run Collective-volitive movement operator
	\ENDWHILE	
\end{algorithmic}

A description of each operator in further details is provided next:

\begin{enumerate}
	
	\item \textbf{Individual movement operator:} every fish in the school performs a local search looking for promising regions . The movement only happens if the new position increases the fish fitness. The position update happen according to Eq. \ref{movInd}.
	
	\begin{equation}
		\label{movInd}
		x_i(t+1)=x_{i}(t)+rand(-1,1)step_{ind},
	\end{equation}
	where $x_{i}(t)$ and $x_{i}(t+1)$ represent the position of the fish $i$ before and after the individual moment operator, respectively. $rand(-1,1)$ is a uniformly distributed random number varying from -1 up to 1 and $step_{ind}$ is a parameter that defines the maximum displacement for this movement.
	
	\item \textbf{Feeding operator: } feeding here means the update of the weight parameter $W_i$ update for fish $i$. This is performed based on the improvement experienced by each fish, which is shown in Eq. \ref{feedOp}.
	
	\begin{equation}
		\label{feedOp}
		W_{i}(t+1)=W_{i}(t)+\frac{\Delta f_i}{max(| \Delta f_i |)},
	\end{equation}
	where $\Delta f_i$ is the fitness variation between the last and the new position and $max(| \Delta f_i |)$ represents the maximum absolute value of the fitness variation among all the fishes in the school. $W$ is only allowed to vary from 1 up to $W_{scale}$, which is a user defined attribute. The weights of all fishes are initialized with the value $W_{scale}/2$.
	
	\item \textbf{Collective-instinctive movement operator: } in this operator, fishes move following the most increasing fitness movements that ocurred in the last individual movement run. An average of the individual movements is calculated based on eq. \ref{Icalc}.
	
	\begin{equation}
		\label{Icalc}
		I=\frac{\sum^{N}_{i=1} \Delta x_{i} \Delta f_{i}}{\sum^{N}_{i=1} \Delta f_{i}}.
	\end{equation}
	and then every fish will be moved according to:
	
	\begin{equation}
		x_i(t+1)=x_{i}(t)+I.
	\end{equation}
	
	\item \textbf{Collective-volitive movement operator: } this operator is employed in order to automatically regulate the exploration/exploitation ability of the school during the search process. The school's barycenter $B$ must be first calculated, which is performed according to Eq. \ref{Bcalc}, taking into account the position $x_{i}$ and the weight $W_{i}$ of each fish:
	
	\begin{equation}
		\label{Bcalc}
		B(t)=\frac{\sum^{N}_{i=1} x_{i}(t) W_{i}(t)}{\sum^{N}_{i=1} W_{i}(t)}.
	\end{equation}
	
	And finally, if the total school weight $\sum^{N}_{i=1} W_{i}$ has increased from the last to the current iteration, the fishes are attracted to the barycenter according to Eq. \ref{volAttrac}. If the total school weight has not improved, fishes are spread away from the barycenter according to Eq. \ref{volSpread}.
	
	\begin{equation}
		\label{volAttrac}
		x_i(t+1)=x_{i}(t)- step_{vol} rand(0,1) * \\ \frac{x_{i}(t) - B(t)}{distance(x_{i}(t),B(t))},
	\end{equation}
	
	\begin{equation}
		\label{volSpread}
		x_i(t+1)=x_{i}(t)+step_{vol} rand(0,1) * \\ \frac{x_{i}(t) - B(t)}{distance(x_{i}(t),B(t))},
	\end{equation}
	where $step_{vol}$ defines the size of the maximum displacement performed with the use of this operator. $distance(x_{i}(t),B(t))$ is the euclidean distance between the fish $i$ position and the school barycenter. $rand(0,1)$ is a uniformly distributed random number varying from 0 up to 1.
	
\end{enumerate}
The parameters $step_{ind}$ and $step_{vol}$ decay linearly according to:

\begin{equation}
	step_{ind}(t+1)=step_{ind}(t)-\frac{step_{ind}(initial)}{It_{max}},
\end{equation}
and similarly:
\begin{equation}
	step_{vol}(t+1)=step_{vol}(t)-\frac{step_{vol}(initial)}{It_{max}},
\end{equation}
where $step_{ind}(initial)$ and $step_{vol}(initial)$ are user defined initial values for $step_{ind}$ and $step_{vol}$, respectively. $It_{max}$ is the maximum number of iterations allowed in the search process.

\subsection{FSS - Stagnation Avoidance Routine}
\label{FSSSAR}

As mentioned before, a modification was proposed in the original FSS in order to make it improve its exploration ability \cite{monteiro2016}. In the original version of the algorithm, the Individual movement component is only allowed to move a fish if it improves the fitness. However, in a very smooth search space, there would be many moving trials with no success and the algorithm could fail to converge.

Further, the Collective-volitive movement operator was designed to automatically regulate the exploration/exploitation ability of the algorithm along the search process. However, in order to do so, this behavior depends on the possibility of the total weight of the school to reduce. If it does not happen, only Eq. \ref{volAttrac} will be utilized in this operator. This means that the ability of attracting the fishes to the school barycenter in order to exploit the search space will always predominate with relation to the ability of spreading the school away from the barycenter in order to allow exploration.

To solve these issues, we introduced a parameter $\alpha$ for which $0 \leq \alpha \leq 1$ in the individual movement operator. $\alpha$ decays exponentially along with the iterations and measures a probability of a worsening allowance for each fish. This means that, every time a fish tries to move to a position that does not improve its fitness, a random number is chosen and if it is smaller than $\alpha$ the movement is allowed. Therefore, only the fishes which presented improvement in their fitnesses within the individual component of the movement can contribute to the $I$ vector calculation used in the collective instinctive movement. In this case, $I$ will be calculated according to:

\begin{equation}
	I=\frac{\sum_{i\in N} \Delta x_{i} \Delta f_{i}}{\sum_{i \in N} \Delta f_{i}},
\end{equation}
where $N$ is the set of all the fishes which improved their fitness in the last Individual movement performed.

This modification is intended to improve the algorithm exploration ability by allowing stochastic worsening movements. However, as the parameter $\alpha$ decays exponentially along the iterations, this effect is intense only in the beginning of the search process and gradually becomes irrelevant in the final of the search.

\subsection{FSS - Not Penalizing Static Success}

FSS-Not Penalizing Static Success (FSS-NPSS) was proposed in the work of Monteiro et al. \cite{Monteiro2016b}. The main modifications  are related to the Feending operator an wirh the inclusion of elitism within the Collective-instinctive movement. The new feeding operator is now defined according to Eq. \ref{weightNPSS}.

\begin{equation}
\label{weightNPSS}
W_{i}(t+1)= 1 + (W_{scale}-1)*\frac{(F_{i}-F_{min})}{(F_{max}-F_{min})},
\end{equation}
where $W_{i}$ and $F{i}$ represent fish $i$ weight and fitness, respectively. $F_{min}$ and $F_{max}$ are the maximum and minimum fitness values found during all the search process.

For the elitism inclusion, fishes which were not able to move within the Individual movement due to the fact that they have already reached a good fitness region can still be able to contribute in the Collective-instinctive movement operator.

In FSS-NPSS, fishes that have not improved will have their displacement $\Delta \textbf{x}_i$ and weight variation $\Delta W_i $ replaced by:

\begin{equation}
\Delta \textbf{x}_{i}^*=step_{vol}\frac{\Delta \textbf{x}_{i}^{t}}{step_{ind}^{t}},
\end{equation}
and:
\begin{equation}
\Delta W_{i}^*=max(\Delta W_{i})\frac{W_{i}-1}{W_{scale}-1},
\end{equation}
where $\Delta \textbf{x}_{i}^*$ and $\Delta W_{i}^*$ are fish $i$ ``fake" displacement and weight increase to be used within the Collective-instinctive movement. $\Delta \textbf{x}_{i}^{t}$ is the last displacement of fish $i$ which occurred in iteration $t$ and $step_{ind}^{t}$ is the step of the Individual movement operator in iteration $t$.

Considering that after the Feeding operator the weight of each fish is a reliable representation of how good is the performance of it during the search, we also propose a modification in the Collective-instinctive movement operator. Instead of using the fitness variation to weight the displacements, this new operator considers weights variation values. Thus, for a school with $N$ fishes, vector $\textbf{I}$ is computed as follows:
\begin{equation}
\textbf{I}=\frac{\sum^{N}_{i=1} \Delta \textbf{x}_{i} \Delta W_{i}}{\sum^{N}_{i=1} \Delta W_{i}}.
\end{equation} 

FSS-NPSS can be combined with FSS-SAR raising other variation known as FSS-NPSS-SAR. This version was chosen to be used in this article in order to provide a comparison on the solution quality provided by the different solution approaches.

\section{Problem Definition}
\label{problemDef}

\subsection{Mixed Model Workplace Time Dependent Assembly Line Balancing Problem}
\label{mmwalbp}

The Mixed Model Workplace Time Dependent Assembly Line Balancing Problem (MMWALBP) was proposed taking simultaneously into account the features of Mixed Model Two Sided and Multi Manned Assembly Line Balancing Problem \cite{Simaria2009a}  \cite{Dimitriadis2006}. Those features are: Mixed models and many workplaces per workstations.

The authors applied the mean model approach, first introduced in the work of \cite{Thomopoulos1967}, in order to tackle the mixed model issue. This means that tasks times are adjusted based on the production level of each model and a joint precedence graph is created. The mean model is utilized to convert a mixed model problem in a single model one.

The main goal of using multi-manned workstations, i.e. workstations containing more than one active workplace, is to minimize the number of workstations of the line while its total effectiveness (in terms of number of workers) remains optimal \cite{Roshani2013}. Multi-manned assembly lines have substantial advantage over a simple assembly line such as reducing: the length of the assembly line and consequently the Work in Progress number of products, the total throughput time, the cost of tools and fixtures, the material handling, workers movement and setup times \cite{Fattahi2011}.

\begin{figure}[htpb]
	\centering
	\caption[Zones definition]{Zones definition}
	\label{workplaces}
	\includegraphics[width=7cm,height=4cm]{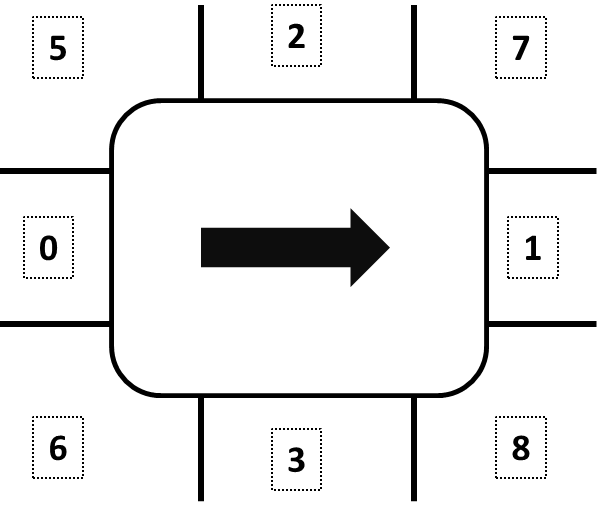}
\end{figure}

MMWALBP has a similar approach when compared to Variable Workplace Assembly Line Balancing Problem (VWALBP) \cite{Becker2009}. The workstation was divided in different zones as it is shown in Figure \ref{workplaces}. The arrow indicates the flow sense of the line. It can be seen that each number identify a work zone within the workstation that will be used to define workplaces. Number 4 represents the zone in the interior of the product to be assembled. Each task contains an attribute telling the decision maker in which zone it should be performed. A maximum number of opened workplaces (operators) per workstation is defined. However, in VWALBP employs a list of prohibited workplaces combinations containing the pairs of workplaces for which a displacement from one to the other would take a relevant amount of time. This was changed within MMWALBP tasks assignment. A time correction procedure was utilized in a way that every displacement between different zones within the workstation generates an additional time in the task that required the displacement.

In order to address the times correction issue, a times matrix must be defined containing the times necessary to move from one workplace to another. The displacement times matrix is specific for each assembly line, hence it should be specifically defined for a real line assembly balancing problem. Many different approaches exist in order to define human operations standard times. One of most common and used approaches is the Method-Times Measurement (MTM) \cite{kanawaty1992introduction}\cite{longo2009effective}.

\subsection{Mixed Model Assembly Line Sequencing Problem}
\label{mmsp}

In Mixed Model Sequencing Problem (MMSP), attention is focused on the shop floor where a decision must be made as to the specific order in which the different models should be launched onto the line. The demand rate of each model may vary. Hence, this short term decision problem must be solved periodically for the appropriate demand rates and the predetermined line balance \cite{Bard1994}.

According to the literature review performed within the work of Boysen et al. \cite{Boysen2009}, MMSP can be grouped according to its optimization target:

\begin{itemize}
	
	\item \textit{Work Overload: }If several models follow each other at the same station, work overloads might occur, which need to be compensated, e.g. by additional utility workers. Work overloads can be avoided if a sequence of models is found, where those models which cause high station times alternate with less work-intensive ones.
	
	\item \textit{Just-in-time-objectives: }An important prerequisite for JIT-supply is a steady demand rate of material over time, as otherwise the advantages of JIT are sapped by enlarged safety stocks that become necessary to avoid stock-outs during demand peaks. Accordingly, JIT-centric sequencing approaches aim at distributing the material requirements evenly over the planning horizon.
	
\end{itemize}

Furthermore, three general sequencing approaches were proposed in literature:

\begin{itemize}
	
	\item Mixed-model sequencing \cite{yano1989survey};
	\item Car sequencing \cite{Solnon2008};
	\item Level scheduling \cite{kubiak1993minimizing};
	
\end{itemize}

In this work, we proposed a FSS based simultaneous solution approach. The balancing version is the aforementioned MMWALBP and the sequencing problem is the Mixed-model sequencing for work overload minimization (MMSP-W). This approach aims at avoiding/minimizing sequence-dependent work overload based on a detailed scheduling which explicitly takes operation times, worker movements, station borders and other operational characteristics of the line into account \cite{Boysen2009}.

In this work, paced straight assembly lines with no parallel workstations and no setup times are taken into account. The workstations are closed. The uncompleted work is performed by utility workers with no concurrency considered.

\section{FSS Solution Procedure}
\label{MMSPPFSS}

\subsection{Balancing and Sequencing Simultaneous Solution Approach}
\label{simBalSeq}

As mentioned before, a simultaneous approach was proposed in order to simultaneously solve both the balancing and sequencing problems of MMALs.

\begin{figure}[htpb]
	\centering
	\caption{Simultaneous solution procedure}
	\label{simultaneousProcedure}
	\includegraphics[width=8cm,height=7cm]{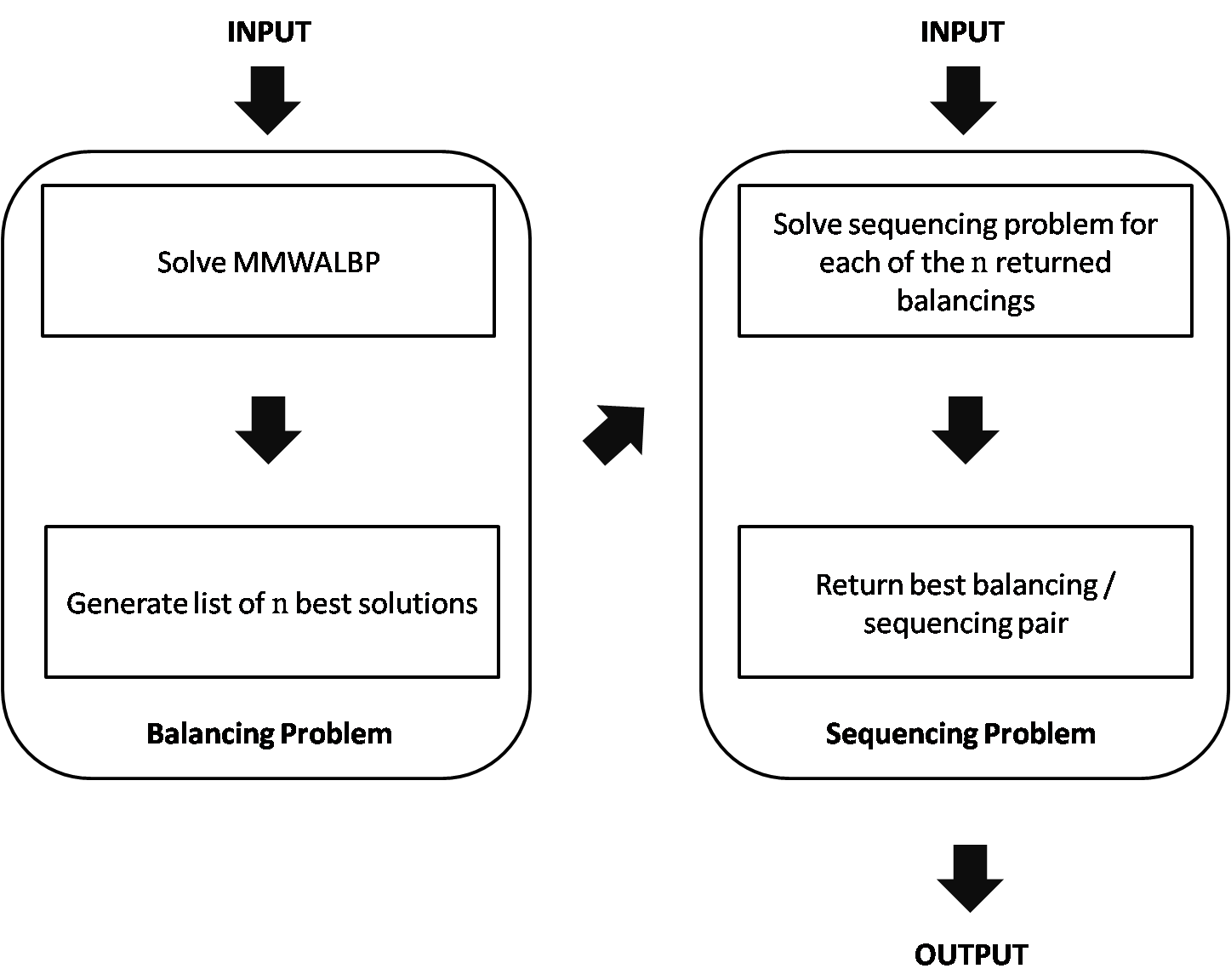}
\end{figure}

The procedure, detailed in the flow shown in Figure \ref{simultaneousProcedure}, is performed according to the following steps:

\begin{enumerate}
	
	\item To input balancing data;
	\item To solve MMWALBP;
	\item Balancing procedures outputs the $n$ (user defined) best solution found;
	\item Input the sequencing data;
	\item To solve the sequencing problem individually for each of the $n$ balancing solutions;
	\item To return the best balancing/sequencing combination found;
	
\end{enumerate}

The solutions representation as well as the objective functions utilized will be described in the following sections.

\subsection{Solution representation}
\label{solutionRepresentation}

Simple procedures were applied in order to map continuous solutions candidates (fishes positions) in discrete valued arrays representing solutions in the problem domain. FSS algorithm was then applied in the continuous space. Once a fitness calculation is required, the fish position is converted into discrete sequences.

\begin{figure}[htpb]
	\centering
	\caption[Random-keys mapping]{Random-keys variation mapping procedure}
	\label{randomkeys}
	\includegraphics[width=5cm,height=3cm]{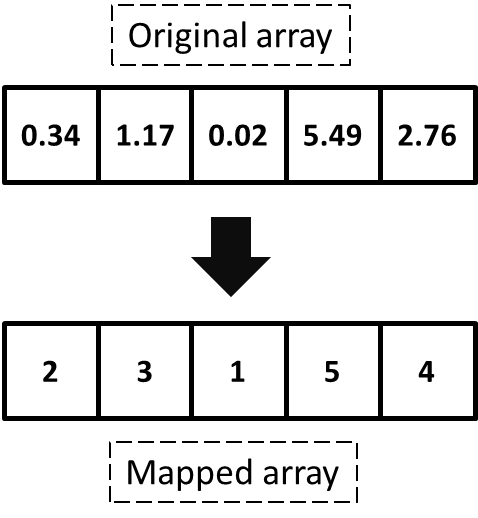}
\end{figure}

Specifically in the case of MMWALBP, the task-oriented approach was applied \cite{scholl1999balancing}. This means that a balancing solution is an permutation of the tasks indexes which is utilized within the assignment procedure for providing fitness calculation. The mapping was performed with a variation of the random-keys \cite{Hamta2013} procedure which converts the smallest value in the fish position array into number 1, the second smallest value into number 2, and this is repeated up to when all the array values are mapped into a task index. Figure \ref{randomkeys} shows an example of an application of the random-keys procedure on a real valued array with five tasks.

For the sequencing portion of the solution, the mapping was proposed through another variation of the aforementioned procedure and it is referred as multiple random-keys. For a given set $i=1, 2, 3...I$ of $I$ different models with production levels given by $P_i$, the total production $P_m$ will be given by $\sum_{i=1}^{I}P_i$. The mapping is performed by assigning $i$ to the $P_I$ smallest values.

The proposed mapping procedure for a given position vector $F$ with dimensions ($P_k$,1) is:

\begin{algorithmic}[1]
	\label{mappingSeq}
	\STATE $A = F$ \COMMENT{Copy $F$}
	\FOR{$i=1$ \TO $I$ } 
	\FOR{$j=1$ \TO $P_i$ }
	\STATE $t=Index(Min(A))$ \COMMENT{$t$ is the index of the minimum value of $A$}
	\STATE $C(t,1)=i$
	\STATE $A(t,1)=\beta$ \COMMENT{$\beta$ is a very large number}		
	\ENDFOR
	\ENDFOR	
	\RETURN $C$
\end{algorithmic}

\subsection{Objective function}
\label{obfunc}

In the balancing portion of the problem solution, it was applied the objective function shown in Eq. \ref{objectiveFunction}.

\begin{equation}
	\label{objectiveFunction}
	minimize \left(K \times \sqrt{\sum_{k=1}^{K}(C-t_{k})^2} \right),
\end{equation}
where $K$ is the number of workplaces (number of operators), $C$ is the cycle time and $t_{k}$ is the workload at workplace $k$.

This modified objective function simultaneously improves the number of workstations and smoothness of the line balancing, furthermore it changes the search space allowing more variation in the fitness when fishes move.

For sequencing problems, many different objective functions appear in literature. Boysen et al. \cite{Boysen2009} cites: minimize work overload \cite{scholl1998pattern}, minimize line length \cite{Bard1994}, minimize throughput time \cite{Bard1992}, minimize maximum displacement of workers from their respective reference point \cite{Okamura1979} among others. Further, in the works of Bautista et al. \cite{Bautista2011} \cite{Bautista2012} the case of workload minimization was considered with the introduction of interruption rules and regularity constraints, respectively.

In this article, we considered the same problem as in the work of Yano and Rachamadugu \cite{Yano1991} for minimization of the total unfinished work or maximization of the total completed work which is the same as minimize work overload \cite{Bautista2008a}. 

In order to compute the objective function, consider: \\

$L$: workstation length (in cycle time units);

$p_{ik}$: process time of job $i$ at workplace $k$;

$s_{ik}$: start time of job $i$ at workplace $k$;

$f_{ik}$: finish time of job $i$ at workplace $k$;

$v_{ik}$: completed work in job $i$ at workplace $k$. \\

Assuming $S_{1k}=0$ for all $k$ workplaces, we have:

\begin{equation}
	s_{ik}=max(i-1,f_{i-1});
\end{equation}
\begin{equation}
	f_{ik}=min(s_{ik}+p_{ik},i-1+L);
\end{equation}
\begin{equation}
	v_{ik}=min(p_{ik},L+i-1-s_{ik});
\end{equation}

The objective function chosen to be maximized, the total completed work, will be given by Eq. \ref{obfuncseq}.

\begin{equation}
	\label{obfuncseq}
	\sum_{i=1}^{P_I}\sum_{k=1}^{K}v_{ik}.
\end{equation}

\section{Experiments and Tests}
\label{experiments}

Tests were performed in six problems derived from SALBP-1 data sets from \url{www.assembly-line-balancing.de}. We have used the sets containing 50 different models with the number of tasks 20, 50 and 100. These problems are referred as $n=20\_50$ (small), $n=50\_50$ (medium) and $n=100\_50$ (large). The first number indicates the quantity of tasks to be allocated and the second stands for the number of different models to be sequenced. In all of these problems, a production plan of 998 models is provided as well as the production level $P_i$ for each model $i$. Cycle time is 1000 in all test cases.

In order to evaluate the performance of FSS variations, more specifically FSS-Vanilla, FSS-SAR and FSS-NPSS-SAR, we compared their outputs against results obtained using PSO. In the PSO version chosen \cite{clerc2002particle}, for a particle $\textbf{x}_i$ its position in iteration $t+1$ is defined in Eq. \ref{psoeq}:

\begin{equation}
	\label{psoeq}
	x_i(t+1) = x_i(t) + \chi [v_{i} + c_1r_1(pb_{i}-x_{i}(t)) + \\ c_2r_{2}(gb_{i} - x_{i}(t))],
\end{equation}
where $\chi = \frac{2}{|2 - (c_1 + c_2) - \sqrt{(c_1 + c_2)((c_1 + c_2) - 4)}|}$ is known as constriction factor and $r_1$ and $r_2$ are uniformly distributed random numbers in the interval $[0; 1]$. In this version $c_1$ and $c_2$ must satisfy $c_1 + c_2 \geq 4$. For this work we have chosen $c_1 = c_2 = 2.1$. The solution of MMWALP with PSO follows the same flow for FSS, as described in Section \ref{MMSPPFSS}. 

FSS, FSS-SAR, FSS-NPSS-SAR and PSO were used to solve all the three aforementioned test instances. We solved the problem instances using the same parameters shown in Table \ref{testParameters}. Each test case was repeated 450 times. In SAR versions of FSS, the value of $\alpha$ parameter in iteration $t$ is $\alpha=0.8e^{-0.007t}$ in all tests performed.

\begin{table}[]
	\centering
	\caption{Test Parameters}
	\label{testParameters}
	\begin{tabular}{cc}
		\hline
		\textbf{$W_{scale}$}  & $10000$                    \\
		\textbf{$Step_{ind}$} & $20\%$                     \\
		\textbf{$Step_{vol}$} & $20\%$                     \\
		\textbf{Search Space} & $\left[-1000; 1000\right]$ \\ \hline
	\end{tabular}
\end{table}

We have applied the widely used Analysis of Variance (ANOVA) technique \cite{winer1971statistical} to establish whether the results obtained are significantly different and, if so, which algorithm is better for each one of the criteria considered.

To apply ANOVA, normality of data should be guaranteed. Thus, 450 outputs for each test case were grouped in samples of size 15 and the means of those were considered as input data for the ANOVA, which results in 30 samples per algorithm for each instance. We applied Shapiro-Wilk test \cite{razali2011power} and then concluded that normality is guaranteed for the results obtained in all test cases.

\subsection{Results}
\label{seqRes}

First of all, a comparison of the four different solution approaches was performed in the three selected test instances in order to evaluate their efficiencies when trying to maximize Completed Work.

Table \ref{balSeqResultsTable} presents the obtained results for different outputs: Completed Work (CW), Workload (WL), number of workplaces (WP) and number of iterations until convergence (IUC), i.e. the last iteration in which an improvement higher than $10^{-4}$ occurred. For these tests, the workstation length was set to $L=0.95$ and the maximum number of workplaces per workstation was set to $3$. The FSS versions are presented only by: Vanilla, SAR and NPSS-SAR.

\begin{table}[]
	\centering
	\caption{Simultaneous Balancing Sequencing results}
	\label{balSeqResultsTable}
	\resizebox{\textwidth}{!}{
	\begin{tabular}{ccccccc}
	\hline
	\textbf{Output} & \textbf{Dataset}        & \textbf{Measure} & \textbf{SAR} & \textbf{NPSS-SAR} & \textbf{PSO} & \textbf{Vanilla} \\ \hline
	& \multirow{2}{*}{Small}  & Mean             & 4496.5119    & 4491.2129         & 4474.8687    & 4493.6283        \\
	&                         & SD               & 28.0606      & 33.1336           & 51.1579      & 31.2959          \\ \cline{2-7} 
	CW              & \multirow{2}{*}{Medium} & Mean             & 3350.5474    & 3356.0287         & 3395.3011    & 3352.2192        \\
	&                         & SD               & 19.9296      & 21.0870           & 32.2869      & 20.5765          \\ \cline{2-7} 
	& \multirow{2}{*}{Large}  & Mean             & 7254.7956    & 7276.9778         & 7148.5438    & 7253.4780        \\
	&                         & SD               & 117.3200     & 141.9520          & 67.1200      & 120.0066         \\ \hline
	& \multirow{2}{*}{Small}  & Mean             & 6.0000       & 6.0000            & 6.0111       & 6.0000           \\
	&                         & SD               & 0.0000       & 0.0000            & 0.1049       & 0.0000           \\ \cline{2-7} 
	WP              & \multirow{2}{*}{Medium} & Mean             & 6.0000       & 6.0000            & 6.0000       & 6.0000           \\
	&                         & SD               & 0.0000       & 0.0000            & 0.0000       & 0.0000           \\ \cline{2-7} 
	& \multirow{2}{*}{Large}  & Mean             & 10.0289      & 10.0844           & 9.1956       & 10.0067          \\
	&                         & SD               & 0.5435       & 0.6345            & 0.4189       & 0.5643           \\ \hline
	& \multirow{2}{*}{Small}  & Mean             & 5575.7109    & 5577.9851         & 5598.0277    & 5574.0965        \\
	&                         & SD               & 20.7721      & 21.1984           & 22.3673      & 22.5772          \\ \cline{2-7} 
	WL              & \multirow{2}{*}{Medium} & Mean             & 3404.2130    & 3406.0236         & 3442.2676    & 3406.2830        \\
	&                         & SD               & 24.9794      & 25.9916           & 34.7513      & 25.1982          \\ \cline{2-7} 
	& \multirow{2}{*}{Large}  & Mean             & 8024.2664    & 8025.5771         & 7985.7387    & 8024.2224        \\
	&                         & SD               & 73.1592      & 74.3165           & 79.2672      & 73.5636          \\ \hline
	& \multirow{2}{*}{Small}  & Mean             & 200.9244     & 231.9178          & 219.2689     & 200.9844         \\
	&                         & SD               & 123.3553     & 126.2045          & 137.5558     & 127.9451         \\ \cline{2-7} 
	IUC             & \multirow{2}{*}{Medium} & Mean             & 267.6311     & 325.3444          & 360.3489     & 274.6689         \\
	&                         & SD               & 145.3887     & 144.2873          & 106.9032     & 147.3617         \\ \cline{2-7} 
	& \multirow{2}{*}{Large}  & Mean             & 245.1978     & 225.5756          & 292.2111     & 241.6711         \\
	&                         & SD               & 0.5435       & 0.6345            & 0.4189       & 0.5643           \\ \hline
\end{tabular}}
\end{table}

An one-way $95\%$ confidence ANOVA was performed in order to evaluate the performance differences among the approaches employed. Calculated degrees of freedom were $v_1 = 2$ and $v_2 = 87$, thus $F_{ref} = 4.89$ in all cases.

Table \ref{fSmallMedium} presents the $F$ statistic provided by ANOVA for Small and Medium data sets. The varying factor was the solution procedure.

\begin{table}[h]
	\centering
	\caption{$F$ values for Small and Medium instances}
	\label{fSmallMedium}
	\begin{tabular}{ccc}
		\hline
		& \textbf{Small} & \textbf{Medium} \\ \hline
		\textbf{CW}  & 27.35          & 373.44          \\ \hline
		\textbf{IUC} & 7.82           & 41.60           \\ \hline
	\end{tabular}
\end{table}

\begin{figure}[h]
	\subfigure[ref1][Completed Work - Small Data Set]{\includegraphics[width=0.47\textwidth]{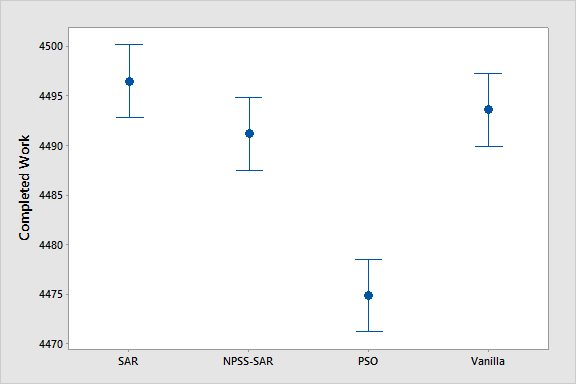}}
	\qquad
	\subfigure[ref2][Completed Work - Medium Data Set]{\includegraphics[width=0.47\textwidth]{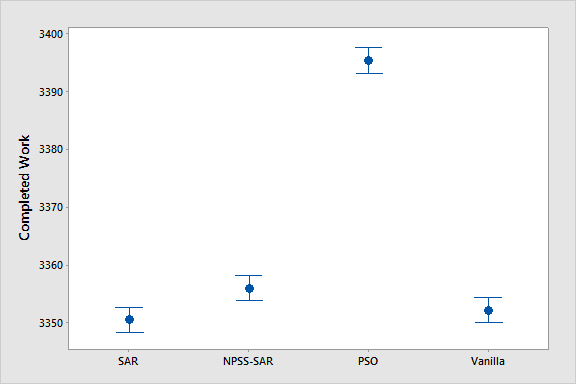}}\\
	\subfigure[ref2][IUC - Small Data Set]{\includegraphics[width=0.47\textwidth]{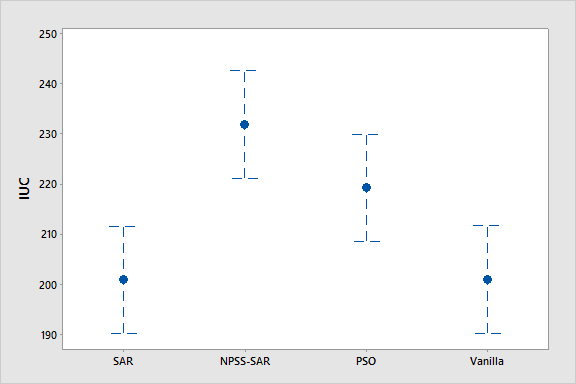}}
	\qquad
	\subfigure[ref2][IUC - Medium Data Set]{\includegraphics[width=0.47\textwidth]{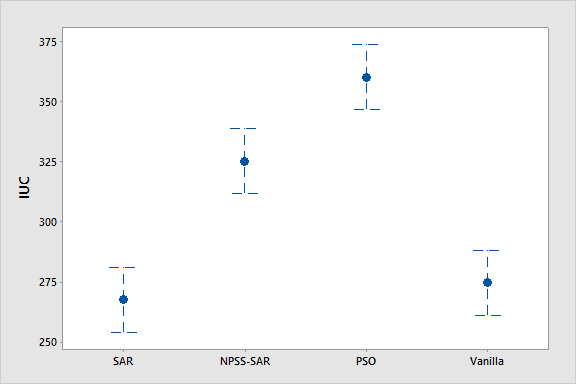}}
	\caption{Confidence intervals based on the pooled standard deviation for the Small and Medium data sets}
	\label{smallMediumCICWIUC}
\end{figure}

$F$ calculated in all test cases considered was greater than $F_{ref}$ which means that we are able to reject the hypothesis that outputs provided by the different approaches do not differ. This can be better illustrated in Figure \ref{smallMediumCICWIUC} where the pooled confidence interval for Completed Work and IUC are presented for the Small and Medium instances.

In the cases of the Completed Work, the overlaps on pooled confidence intervals in either Small and Medium instances suggest the data compared is likely to come from the same probability distribution and do not differ significantly. However, PSO algorithm presented a lower output in Small case and a best result in the Medium problem. In terms of convergence speed, IUC for FSS-NPSS-SAR was always higher than the other approaches, which means that this version usually takes longer to converge with no relevant output improvement.

Specifically for the Large data set, a more detailed analysis was carried out and ANOVA was applied for all the outputs available. Table \ref{fLarge} presents the $F$ statistic values for: Completed Work (CW), number of workplaces (WP), the ratio Completed Work/Workload (CW/WL) and the number of iterations until convergence (IUC). Once more, the test performed suggest that at least one of the approaches employed differs of the others for all the outputs considered.

\begin{table}[h]
	\centering
	\caption{$F$ values for Large instance}
	\label{fLarge}
	\begin{tabular}{cc}
		\hline
		\textbf{Output} & \textbf{$F$} \\ \hline
		\textbf{CW}     & 102.32       \\
		\textbf{IUC}    & 18.88        \\
		\textbf{WP}     & 273.31       \\
		\textbf{CW/WL}  & 57.54        \\ \hline
	\end{tabular}
\end{table}

\begin{figure}[h]
	\subfigure[ref1][Completed Work]{\includegraphics[width=0.47\textwidth]{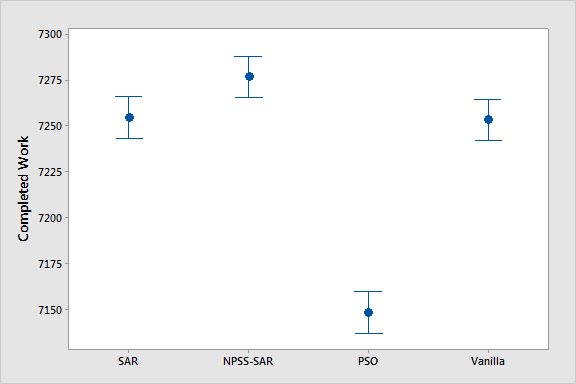}}
	\qquad
	\subfigure[ref2][Workload]{\includegraphics[width=0.47\textwidth]{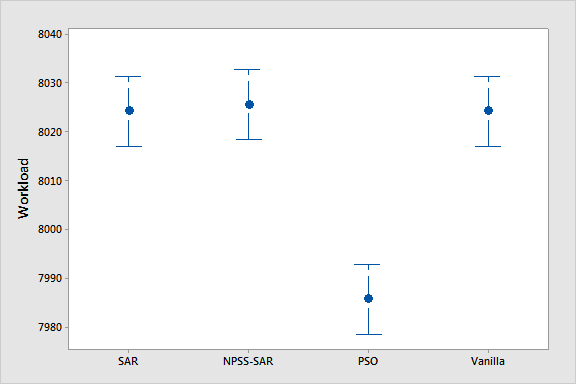}}\\
	\subfigure[ref2][CW/WL ratio]{\includegraphics[width=0.47\textwidth]{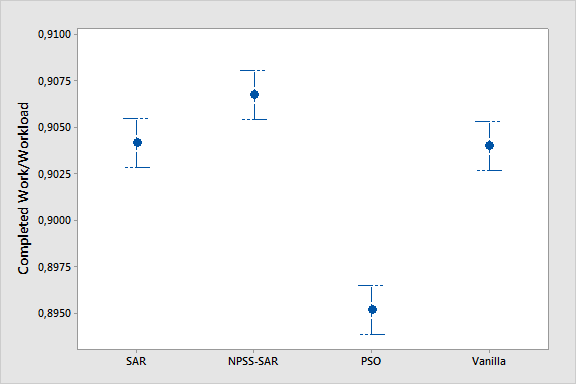}}
	\qquad
	\subfigure[ref2][Iterations Until Convergence]{\includegraphics[width=0.47\textwidth]{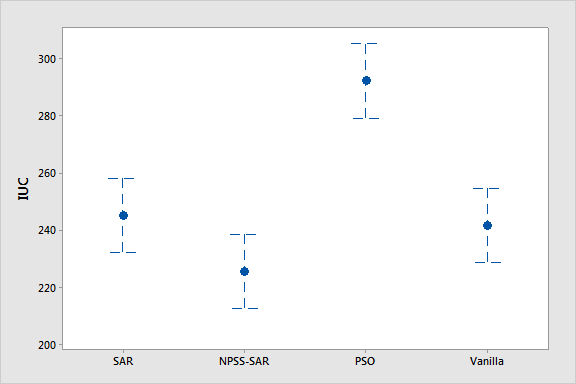}}
	\caption{Confidence intervals based on the pooled standard deviation for the Large data set}
	\label{largeCICWIUC}
\end{figure}

The aforementioned can be better analyzed with basis in Figure \ref{largeCICWIUC}. The pooled confidence intervals for WP indicate that PSO was able to find balancing outputs better than those returned by FSS variations. However, the results of the ratio Completed Work/Workload make clear that a higher fraction of the workload is completed in the outputs provided by FSS-NPSS-SAR. This means that the balancing/sequencing solution provided by PSO seems more efficient once it utilizes less workplaces, but in the other hand this solution requires more utility work (or line stoppages) in order to finish the production plan. The higher number of workplaces of FSS-NPSS-SAR increases the workload but also provides a higher proportion of completed work requiring less utility work or reducing the risk of line stoppages. For IUC, PSO took longer to converge in the Large instance.

Furthermore, a comparison was performed in order to study the effect of the variation of the workstation maximum permitted number of workplaces per workstation. Values $2$, $4$ and $6$ were employed as independent variables in an ANOVA using large data set and FSS-SAR as solution approach. Results are presented in Table \ref{resultsnVariando} and $F$ values are shown in Table \ref{fnVariando}.

\begin{table}[h]
	\centering
	\caption{Simultaneous Balancing and Sequencing results with the variation of the maximum number of workplaces per workstation}
	\label{resultsnVariando}
	\begin{tabular}{ccccc}
	\hline
	\textbf{}                       & \textbf{} & \textbf{2} & \textbf{4} & \textbf{6} \\ \hline
	\multirow{2}{*}{\textbf{WP}}    & Mean      & 10.0000    & 11.9844    & 12.0000    \\
	& SD        & 0.0000     & 0.1239     & 0.0000     \\ \hline
	\multirow{2}{*}{\textbf{WL}}    & Mean      & 8273.4059  & 8121.8921  & 7964.2821  \\
	& SD        & 50.5872    & 78.2892    & 59.3590    \\ \hline
	\multirow{2}{*}{\textbf{CW/WL}} & Mean      & 0.9194     & 0.9612     & 0.9577     \\
	& SD        & 0.0055     & 0.0062     & 0.0053     \\ \hline
	\multirow{2}{*}{\textbf{CW}}    & Mean      & 7606.7499  & 7806.5702  & 7627.5828  \\
	& SD        & 29.5040    & 61.8566    & 52.6124    \\ \hline
	\end{tabular}
\end{table}

\begin{table}[h]
	\centering
	\caption{$F$ values for the variation of the maximum number of workplaces per workstation}
	\label{fnVariando}
	\begin{tabular}{cc}
		\hline
		\textbf{Output} & \textbf{$F$} \\ \hline
		\textbf{CW}     & 2561.24      \\
		\textbf{CW/WP}  & 6289.95      \\ \hline
	\end{tabular}
\end{table}

\begin{figure}[H]
	\subfigure[ref1][Completed Work]{\includegraphics[width=0.47\textwidth]{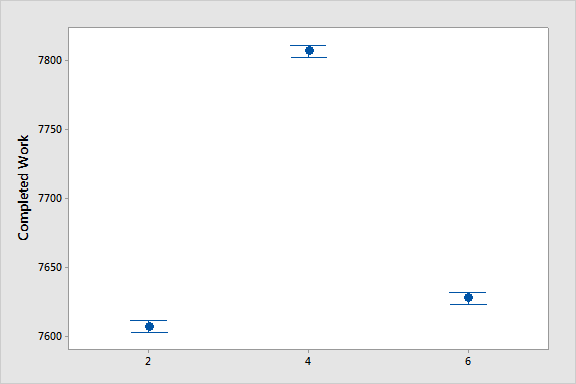}}
	\qquad
	\subfigure[ref2][CW/WL ratio]{\includegraphics[width=0.47\textwidth]{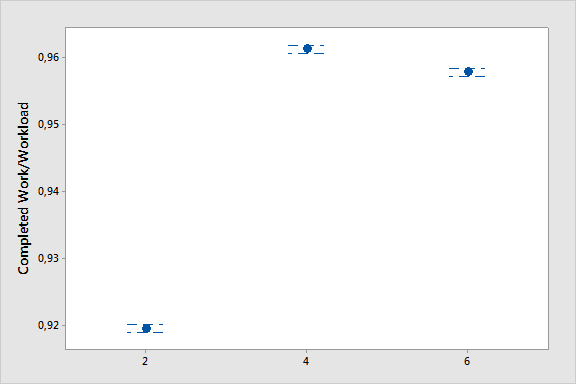}}
	\caption{Confidence intervals based on the pooled standard deviation for the variation of the maximum number of workplaces per workstation}
	\label{nVariandoCICWIUC}
\end{figure}

The pooled confidence intervals are presented in Figure \ref{nVariandoCICWIUC}. The case in which the maximum number of workplaces is $4$ is the one in which the best ration CW/WL occurred due to the excess of displacements in cases where few workplaces are allowed and the excess of precedence relation conflicts in the case which several workplaces are allowed.

\newpage
\section{Conclusion}
\label{conclusion}

In this work we have applied an nature-inspired metaheuristic called Fish School Search algorithm in order to solve two relevant decision problems: Assembly Line Balancing Problem and Mixed Model Assembly Line Sequencing Problem. The relevance of the tackled problems relies on their NP-Hard nature as well as in practitioners interest within Industry.

FSS is a continuous optimization technique and both of aforementioned problems have combinatorial nature. Hence, mapping procedures were applied in order to convert fishes positions into sequences of tasks for the balancing portion of the solution approach and a sequence of models for the sequencing part.

Furthermore, both problems are known to be strongly interdependent. Thus, a simultaneous approach was applied in a sense that first the balancing problem is solved and a list of the best solutions found within the search are used as input for the sequencing problem solution. The best combination of balancing/sequencing are returned as solution of the problem.

Tests were performed in three different data sets and a comparison between the original version of FSS and two variations referred as FSS-SAR and FSS-NPSS-SAR as well as Particle Swarm Optimization algorithm was carried out. Results show that, in general, FSS versions generate solution requiring more workplaces than PSO. However, this fact ends up resulting in more efficient results for the sequencing portion once the outputs of FSS require less use of utility work. FSS-NPSS-SAR was the variation of FSS which returned the best results.

As future work, some niching able metaheuristic \cite{buarque2014weight,Madeiro2011} may be applied in the balancing portion of the solution approach. Hence, the list containing balancing best found solutions would be directly the output of the multi-solution procedure. Moreover, different combinations of the approaches applied in each portion of the solution (balancing/sequencing) can be applied for performance evaluation.

\newpage

\bibliographystyle{abbrv}
\bibliography{bibliografia}

\end{document}